%% file: main.tex
\input{preamble}

\begin{document}

\title{{Physics-Informed Neural Networks for Biological $2\mathrm{D}{+}t$ Reaction--Diffusion Systems}
\thanks{
JAC and SH were supported by the Kjell och M{\"a}rta Beijer Foundation.
SH was additionally supported by Wenner-Gren Stiftelserna/the Wenner-Gren Foundations (WGF2022-0044), and the Swedish Research Council (project 2024-05621). 
DST was supported by the Research Council of Norway (grant 325628/ IAR). DST gratefully acknowledges Professor Stefan Krauss for providing access to the imaging platform at the Hybrid Technology Hub, University of Oslo.
\textit{1: DST second affiliation: Dept. of Translational Hematology and Oncology Research, Cleveland Clinic, Cleveland, USA.}
}}

\input{author_block}

\maketitle

\begin{abstract}
\input{abstract}

\end{abstract}


\section{Introduction}
\input{introduction}

\section{Methods} \label{sec:methods}
\input{Methods/Eq_architecture_MLPdesign_TV_ES}

\input{Methods/loss_functions}

\input{Methods/SR}

\section{Results} \label{sec:Experimental_Results}
\input{Methods/experimental_data_pipeline}
\input{Results/results}

\vspace{0.1cm}
\section{Conclusions and Future Work} \label{sec:Conclusion}
\input{future_work}

\bibliographystyle{IEEEtran}
\bibliography{references}

\appendices 

\section{}
\label{appendix1}
\input{Appendices/A1}
\section{}
\label{appendix2}
\input{Appendices/A2}

\end{document}

%% file: preamble.tex
\documentclass[conference]{IEEEtran}
\IEEEoverridecommandlockouts
\usepackage{cite}
\usepackage{amsmath,amssymb,amsfonts}
\usepackage{algorithmic}
\usepackage{graphicx}
\usepackage{textcomp}
\usepackage{xcolor}
\def\BibTeX{{\rm B\kern-.05em{\sc i\kern-.025em b}\kern-.08em
    T\kern-.1667em\lower.7ex\hbox{E}\kern-.125emX}}

\usepackage[caption=false,font=footnotesize]{subfig}

\usepackage{booktabs}
\usepackage{tabularx}
\usepackage{array}

\usepackage{amsmath}     
\usepackage{tikz}
\usepackage{hyperref}

\newcommand{\cornerlabelimgshift}[5][]{%
  \begin{tikzpicture}
    \node[inner sep=0] (img) {\includegraphics[#1]{#2}};
    \node[
      anchor=#4,
      font=\bfseries\small,
      fill=white,
      fill opacity=0,
      text opacity=1,
      inner sep=1pt,
      #5
    ] at (img.#4) {#3};
  \end{tikzpicture}%
}


%% file: author_block.tex
\newcommand{\thirdwidthauthor}[1]{\parbox{0.325\textwidth}{\centering #1}}

\author{\IEEEauthorblockN{William Lavery}
\IEEEauthorblockA{\thirdwidthauthor{\textit{Dept. of Information Technology}} \\
\textit{Uppsala University}\\
Uppsala, Sweden \\
william.lavery@it.uu.se}
\and
\IEEEauthorblockN{Jodie A. Cochrane}
\IEEEauthorblockA{\thirdwidthauthor{\textit{Dept. of Information Technology}} \\
\textit{Uppsala University}\\
Uppsala, Sweden \\
jodie.cochrane@it.uu.se}
\and
\IEEEauthorblockN{Christian Olesen}
\IEEEauthorblockA{\thirdwidthauthor{\textit{Dept. of Medical Genetics}} \\
\textit{Oslo University Hospital}\\
Oslo, Norway \\
chole4872@oslomet.no}
\and
\IEEEauthorblockN{Dagim S. Tadele$^1$}
\IEEEauthorblockA{\thirdwidthauthor{\textit{Dept. of Medical Genetics}}\\
\textit{Oslo University Hospital}\\
Oslo, Norway \\
dagtad@ous-hf.no}
\and
\IEEEauthorblockN{John T. Nardini}
\IEEEauthorblockA{\thirdwidthauthor{\textit{Dept. of Mathematics and Statistics}} \\
\textit{The College of New Jersey}\\
Ewing, USA \\
nardinij@tcnj.edu}
\and
\IEEEauthorblockN{Sara Hamis}
\IEEEauthorblockA{\thirdwidthauthor{\textit{Dept. of Information Technology}} \\
\textit{Uppsala University}\\
Uppsala, Sweden \\
sara.hamis@it.uu.se}
}

%% file: abstract.tex
Physics-informed neural networks (PINNs) provide a powerful framework for learning governing equations of dynamical systems from data. Biologically-informed neural networks (BINNs) are a variant of PINNs that preserve the known differential operator structure (e.g., reaction--diffusion) while learning constitutive terms via trainable neural subnetworks, enforced through soft residual penalties. Existing BINN studies are limited to \(1\mathrm{D}{+}t\) reaction--diffusion systems and focus on forward prediction, using the governing partial differential equation  as a regulariser rather than an explicit identification target. Here, we extend BINNs to \(2\mathrm{D}{+}t\) systems within a PINN framework that combines data preprocessing, BINN-based equation learning, and symbolic regression post-processing for closed-form equation discovery.
We demonstrate the framework’s real-world applicability by learning the governing equations of lung cancer cell population dynamics from time-lapse microscopy data,
recovering \(2\mathrm{D}{+}t\) reaction--diffusion models from experimental observations. The proposed framework is readily applicable to other spatio-temporal systems, providing a practical and interpretable tool for fast analytic equation discovery from data.

%% file: introduction.tex
Modelling the dynamics of biological systems has a long and rich history drawing on diverse mathematical techniques. Particularly, many formulations are based on ordinary and partial differential equations (ODEs and PDEs) derived from first-principles kinetics or biological theory \cite{Allen2007}. 
Although these approaches have been profoundly effective in capturing  the behaviour of biological systems on various scales--from molecular \cite{CornishBowden2013} to epidemiological modelling \cite{Brauer2019}--many biological processes remain poorly characterised mathematically.

Modern data platforms are providing insights into such processes through high-dimensional, longitudinal data streams from single-cell multi-omics \cite{Yu2025}, time-lapse microscopy \cite{Wang2025}, wearable biosensors \cite{Clausen2025,Spinelli2025}, and environmental metagenomics \cite{Becsei2024}, 
which implicitly contain information about the system dynamics. Combined with advances in high-performance computing, this has accelerated the development of equation learning (EQL), a family of methods that seek to infer closed-form dynamical rules from temporal observations  \cite{campsvalls2023discoveringcausalrelationsequations}. Formally, in EQL, empirical data are assumed to originate from an unknown system for a quantity of interest \(u(\mathbf{x},t)\) evolving according to  
\begin{equation}
\frac{\partial u(\mathbf{x},t)}{\partial t} \;=\; \mathcal{F}(\cdot),
\label{eq:generic_dynamical_system}
\end{equation}
with $d$-dimensional spatial coordinates \(\mathbf{x}\in\Omega\subseteq\mathbb{R}^{d}\) and time variable \(t\in[t_0,t_f]\). 
The operator \(\mathcal{F}\) may include explicit dependence on space, time, the field itself, and its spatial derivatives. As such, EQL algorithms aim to uncover the functional form of \(\mathcal{F}\) from data without fully specifying the right-hand side of \eqref{eq:generic_dynamical_system}. EQL thus 
lies at the intersection of mechanistic modelling and modern machine learning \cite{Baker2018}, offering a data-driven route to discover compact, interpretable dynamical equations for systems where traditional derivations may falter. Three methodological paradigms currently dominate the EQL field: 
\vspace{.2cm}

\noindent\textbf{i. Sparse-regression approaches} formulate the relationship between $\frac{\partial u}{\partial t}$ and some user-specified library of non-linear candidate terms as a linear regression problem. In standard formulations, filtering or spline techniques denoise the data, and derivatives of $u$ are approximated 
to construct the library. Weak formulations instead employ integral projections to reduce noise sensitivity by transferring derivatives to smooth test functions \cite{Messenger2021WSINDy, Messenger2021WSINDyPDE}. Once 
a linear system is constructed by either method, sparsity-promoting optimisation selects a minimal subset of the library that best explains the data \cite{Brunton2016,Rudy2017,Wei2022,Fung2025}. This mitigates the need for an exhaustive combinatorial search, but constrains the discovered dynamics, $\mathcal{F}$, to the span of the chosen library. 
As an extension, hybrid neural sparse methods combine sparse regression approaches with neural function approximation. 
These methods use deep neural networks (DNNs) to learn feature representations from which candidate terms are constructed and then selected via sparse regression \cite{Champion2019,Chen2021,Stephany2024,Gao2025}. This reduces reliance on a user-specified candidate library by leveraging the expressive capacity of neural surrogate functions, but may introduce non-physical or redundant features and increase sensitivity to model selection.
\\[1em]
\noindent\textbf{ii. Physics-informed neural networks (PINNs)} use a DNN to approximate the solution $u$ in \eqref{eq:generic_dynamical_system} by enforcing a mechanistic equation during training. Since the introduction of PINNs \cite{Raissi2019}, the methodology  has diversified immensely. For example, Bayesian PINNs extend the formulation by learning distributions over the network parameters and solution fields rather than a single deterministic approximation, improving robustness when observations are noisy \cite{Liu2021}.
Furthermore, a range of architectures for hard-constraining \eqref{eq:generic_dynamical_system} have emerged that incorporate finite-difference stencils or equality constraints directly into the DNN, thereby preventing the accumulation of PDE or boundary constraint violations and maintaining physically consistent predictions in long-term simulations \cite{Jagtap2020,Liu2024,Li2024ops}. However, the mechanistic form of the governing PDE operator is typically assumed to be known \textit{a priori}, with learning restricted to its parameters. This assumption is relaxed in universal PINNs (UPINNs), which also learn missing or unknown components of the operator \cite{Podina2024UPINN}. %
\\[1em]
\noindent\textbf{iii. Neural differential equation approaches} enforce dynamics through forward solving differential equations. 
A prominent case is universal differential equations (UDEs), which embed trainable neural components directly within differential equation solvers, partitioning the dynamics into known and unknown components \cite{rackauckas2021universaldifferentialequationsscientific}. This enables known physics to be enforced through the numerical ODE/PDE solver while the neural networks learn latent dynamics that are unknown or not captured by the mechanistic model. However, UDEs require repeated differential equation simulation during training which can lead to high computational cost and optimisation challenges.
In the limiting case of fully data-driven modelling, neural ODEs parameterise $\mathcal{F}$ entirely with a neural network, at the cost of reduced interpretability and greater dependence on data quality and coverage \cite{neuralODE}.\\[1em]
Methodologies i-iii differ in how much of $\mathcal{F}$ is specified \textit{a priori} and in how constraints are imposed. Biologically-informed neural networks (BINNs) occupy a niche of UPINNs, preserving coarse PDE structure while learning constitutive components as neural subnetworks, making them well suited to biological systems where mechanisms are partially known and data are noisy and sparse \cite{Lagergren2020}. 
This study extends prior BINN methodology. 
Specifically, we:

\begin{enumerate}
\item Generalise the BINN architecture to $2\mathrm{D}{+}t$.
\item Develop a PINN framework that augments the BINN architecture with symbolic regression (SR) post-processing to convert neural network outputs into analytical functions.
\item Apply the PINN framework 
to learn $2\mathrm{D}{+}t$ reaction--diffusion dynamics of lung cancer cell populations from time-lapse microscopy data.
\end{enumerate}
\vspace{.1cm}

The PINN framework developed in this study is provided in an open access \href{https://github.com/williamlavery/pinn-reaction-diffusion-2dt.git}{GitHub repository} containing a user-friendly, interactive notebook, and can be readily adapted to custom reaction--diffusion systems and other systems governed by similar PDEs.

%% file: Methods/Eq_architecture_MLPdesign_TV_ES.tex
\subsection{The governing reaction--diffusion equation}
In this study, we work with reaction--diffusion equations to describe how biological systems evolve in time and space. Specialising \eqref{eq:generic_dynamical_system} to our setting, the dynamics of the quantity $u(\mathbf{x},t)$ are governed by 
\begin{align}
\begin{split}
    \frac{\partial u(\mathbf{x},t)}{\partial t} ={} & \nabla \cdot \left[D(u(\mathbf{x},t)) ~ \nabla u(\mathbf{x},t) \right] \\
    & \quad \qquad + G(u(\mathbf{x},t)) ~ u(\mathbf{x},t),
\end{split}
\label{eq:reaction_diff_eq_2d}
\end{align}
where $D$ is a function describing diffusion, and $G$ is a function for (per-capita) growth. The EQL problem becomes learning the diffusion and growth functions from $u$-data. 

\subsection{The BINN architecture} \label{sec:BINN_architecture}
The BINN architecture comprises three fully connected neural networks, also known as multilayer perceptrons (MLPs): (i) $\mathrm{NN}_u(\mathbf{x},t;\theta_u): \mathbb{R}^{d+1}\to\mathbb{R}$, which maps spatiotemporal inputs to predicted densities $\hat u(\mathbf{x},t)$; 
(ii) $\mathrm{NN}_D(\hat u;\theta_D): \mathbb{R}\to\mathbb{R}$, which maps $\hat u$ to diffusion predictions $\hat D(\hat u)$; 
and (iii) $\mathrm{NN}_G(\hat u;\theta_G): \mathbb{R}\to\mathbb{R}$, which maps $\hat u$ to growth predictions $\hat G(\hat u)$. Here, $\hat u$, $\hat D$, and $\hat G$ denote neural network surrogates of the functions $u$, $D$, and $G$, in~\eqref{eq:reaction_diff_eq_2d}. 
All network parameters $\{\theta_u,\theta_D,\theta_G\}$ are jointly optimised 
to fit the data while satisfying the PDE constraints and optional biological restrictions.
The general architecture is illustrated in Fig.~\ref{fig:fig1}, and we here follow the BINN design choices and training protocols discussed in our recent study \cite{lavery2026binn}. These choices and protocols are summarised below.
\input{Methods/figs/fig1}

\vspace{.1cm}
\subsubsection{MLP design}
Each MLP consists of three hidden layers of equal width followed by a single-neuron output layer, providing sufficient depth to capture nonlinear relationships while maintaining a compact architecture. All hidden layers use SiLU activations. The density MLP $\mathrm{NN}_u$ uses wider hidden layers each containing $64$ neurons, while the diffusion and growth MLPs use narrower layers of $4$ neurons each. This choice reflects the assumption that $D$ and $G$ are relatively simple functions, whereas $u$ is more complex. Output activations are selected to enforce physical constraints: softplus is used for $\mathrm{NN}_u$ and $\mathrm{NN}_D$ to ensure non-negativity of density and diffusivity, while $\mathrm{NN}_G$ uses a linear output, allowing both positive and negative population growth.
\vspace{.1cm}

\subsubsection{The training--validation (TV) split}
In the BINN architecture, density data are randomly split into training and validation sets. 
Specifically, data are discretised into a tensor 
$u_{\text{data}} \in \mathbb{R}^{n_{x_1}\times n_{x_2}\times n_t}$ for $2\text{D}{+}t$ systems, where $n_{x_1}, n_{x_2}$ denote the fixed number of bins in each spatial direction, and $n_t$ the number of observed time points. The entries of $u_{\text{data}}$ are then 
randomly split into training and validation sets by assigning grid points independently 
of space and time, and this partition is fixed throughout training. The training set is used for network parameter updates, whereas the validation set is used to implement early stopping (ES) \cite{Lagergren2020}. Both the training and validation sets are subsequently used for final model selection in post-training analysis. In this analysis, the learned diffusion and growth MLPs are evaluated over densities from the span of the combined training and validation sets, with the aim of accurately capturing biological mechanisms within the observed $u$-data range rather than extrapolating beyond it. 
We use five random 80/20 TV splits to form a small ensemble and assess sensitivity to data partitioning in this low-data regime, as data partitioning influences both model performance and interpretation.
\vspace{.1cm}
\subsubsection{Early stopping} \label{sec:ES}
We use an ES protocol to balance convergence of the training objective with computational efficiency. 
In particular, training is terminated when the validation loss fails to improve by at least 5\% for a fixed number of consecutive epochs, corresponding to the ES patience, i.e., the maximum number of epochs without improvement before training is halted. 
The model parameters from the last epoch satisfying this improvement criterion are selected as the trained parameters.

%% file: Methods/figs/fig1.tex
\begin{figure}[t]
    \centering
    \begin{minipage}[c]{0.45\textwidth}
        \centering
            \includegraphics[width=\linewidth]{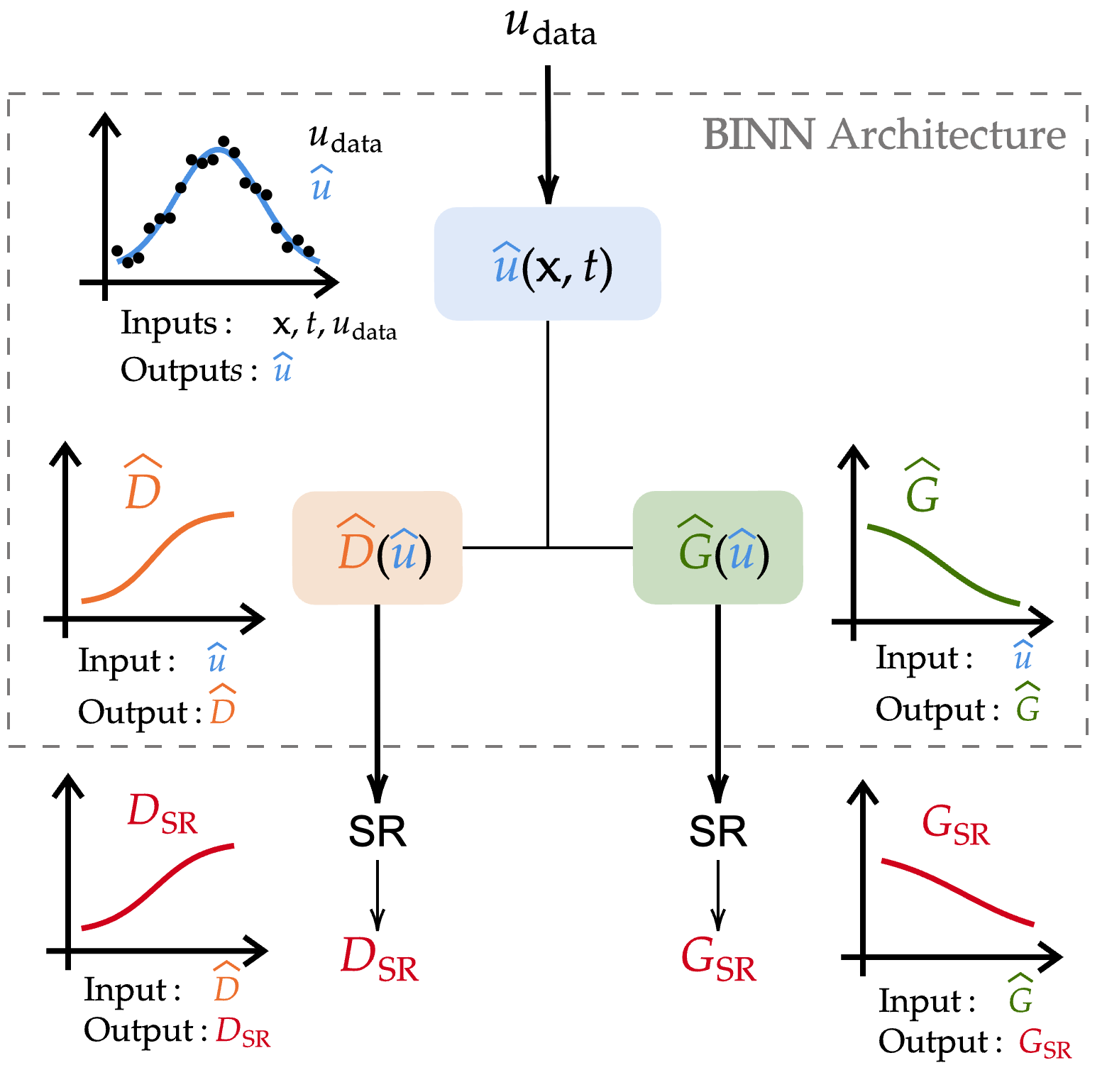}
    \vspace{0cm}
    \end{minipage}
    \hfill
    \begin{minipage}[c]{0.48\textwidth}
        \centering
        \includegraphics[width=\linewidth]{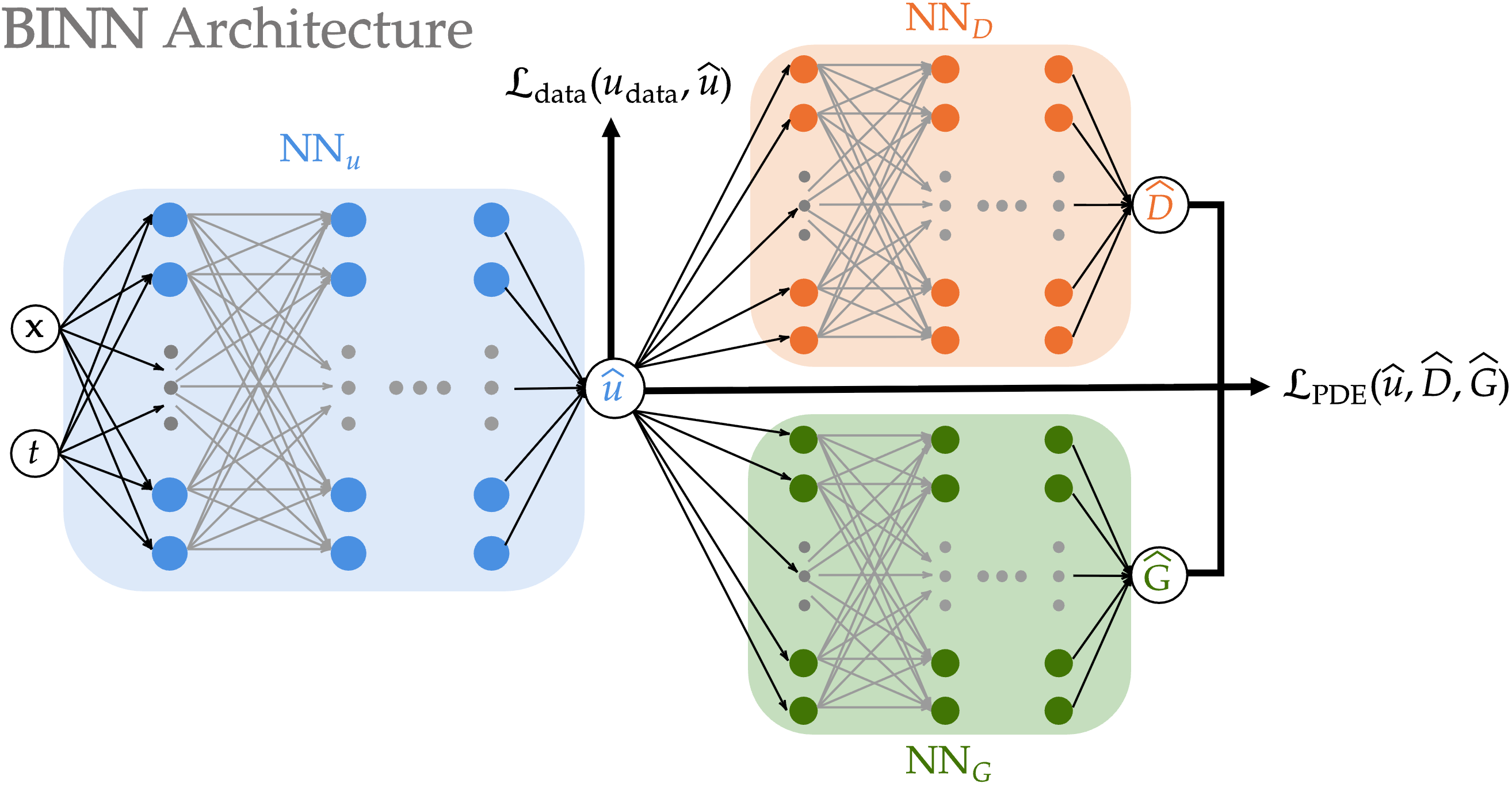}
    \end{minipage}
    \caption{PINN framework developed in this work (top) with the BINN architecture highlighted (bottom).}

\label{fig:fig1}
\end{figure}

%% file: Methods/loss_functions.tex
\subsection{The loss functions} \label{sec:losses}

In BINNs the networks $\mathrm{NN}_u$, $\mathrm{NN}_D$, and $\mathrm{NN}_G$ are jointly trained by minimising the total loss function
\begin{align}
\mathcal{L}_{\text{Total}}
=
\lambda_{\text{data}}\,\mathcal{L}_{\text{data}}
+
\lambda_{\text{PDE}}\,\mathcal{L}_{\text{PDE}},
+
\lambda_{\text{bio}}\,\mathcal{L}_{\text{bio}},
\label{eq:total_loss}
\end{align}
where $\lambda_{\text{data}},\lambda_{\text{PDE}},\lambda_{\text{bio}}>0$ weight the data, PDE, and biological losses respectively.  All quantities in the loss terms are normalised such that the data and PDE residuals are of comparable magnitude at the start of training, allowing us to set $\lambda_{\text{data}}=\lambda_{\text{PDE}}=1$ \cite{Lagergren2020}. This choice is maintained throughout training to avoid the increased computational costs of optimal dynamic weighting strategies. In our experimental setting, 
there is limited information restricting the functional forms of $D$ and $G$; 
accordingly, we set $\lambda_{\text{bio}} \equiv 0$. Details on the data and PDE loss terms are provided below. 
\vspace{.1cm}
\subsubsection{Data loss}
The data loss enforces agreement between the model predictions and the observed data. 
We assume observations are corrupted by additive Gaussian noise with non-constant variance, consistent with previous cell migration modelling studies \cite{Lagergren2020,Lagergren2020_b}. 
Specifically, for a system with two spatial dimensions, 
\begin{align}
u_{\text{data}}^{i,j,s}
=
u(x_{1,i},x_{2,j},t_s)
+
\omega_{i,j,s} \, u^\gamma(x_{1,i},x_{2,j},t_s)\,\varepsilon_{i,j,s}.
\label{eq:noise_model}
\end{align}

Here, $i,j$ index spatial locations and $s$ indexes time; $u(\cdot)$ denotes the PDE solution for the cell density; $\varepsilon_{i,j,s} \sim \mathcal{N}(0,1)$ are independent Gaussian draws; $\gamma \in \mathbb{R}$ controls how the noise magnitude scales with density; and $\omega_{i,j,s} \in \mathbb{R}$ are additional variance scaling coefficients that we, for simplicity, set to be constant.
The noise model \eqref{eq:noise_model} produces a generalised least-squares loss. However following \cite{lavery2026binn}, to ensure robustness in low-density regions and consistent optimisation, we set $\gamma=0$, producing the ordinary least squares objective function,
\begin{align}
\mathcal{L}_{\text{data}}
=
\frac{1}{n_{x_1}n_{x_2}n_t}
\sum_{i,j,s}
\big[
u_{\text{data}}^{i,j,s}
-
\hat{u}(x_{1,i},x_{2,j},t_s)
\big]^2.
\end{align}

\subsubsection{PDE loss}
The governing PDE is enforced by comparing the two sides of \eqref{eq:reaction_diff_eq_2d} at a set of $n_c$ randomly sampled collocation points 
$\{(x_{1,k}, x_{2,k}, t_k)\}_{k=1}^{n_c}$ 
drawn uniformly over the spatio–temporal data domain. Derivatives are computed via automatic differentiation, yielding the residual loss
\begin{align}
\mathcal{L}_{\text{PDE}}
=
\frac{1}{n_c}
\sum_{k=1}^{n_c}
\left[
\left.\frac{\partial \hat{u}}{\partial t}\right|_{k}
-
\left.\nabla\!\cdot\!\left(\hat{D}\,\nabla \hat{u}\right)\right|_{k}
-
\left.\hat{G}\,\hat{u}\right|_{k}
\right]^2,
\end{align}
where %
$\left.\cdot\right|_k$  
denotes evaluation at the $k$-th collocation point.

%% file: Methods/SR.tex
\subsection{Symbolic Regression} \label{sec:SR}
To convert MLP outputs into interpretable analytic expressions we employ SR, implemented via the \texttt{PySR} package \cite{Cranmer2023}. This method uses an evolutionary algorithm and is chosen for its flexibility, suitability for scientific applications, and explicit control over expression complexity. The considered \texttt{PySR} hyperparameters are selected to balance exploration, biological plausibility, and computational cost (Table~\ref{tab:Methods:secSR:PySR_params}). 
As outlined in 1-3 below, SR is applied multiple times to the constructed ensemble diffusion and growth predictions to generate candidate symbolic expressions. To ensure robustness, dominant expressions are then selected from these candidates to define $D_{\mathrm{SR}}$ and $G_{\mathrm{SR}}$.

\input{Methods/Tables/SR}
\input{Methods/figs/fig_results_schematic}

\vspace{.1cm}
\subsubsection{Constructing ensemble predictions} 
We focus on well-supported densities to construct the ensemble predictions. Namely, we use the intersection of the central 90\% intervals of the training $u$-data across all TV splits. This produces conservative bounds with shared density support. Within these bounds, the diffusion and growth MLP predictions are weighted in proportion to the replicate- and TV-split-specific training density distribution. The weighted predictions are then averaged across TV splits to produce ensemble diffusion and growth predictions.

\vspace{.1cm}
\subsubsection{Generating candidate expressions}
SR is applied multiple times to the ensemble predictions using different random seeds to account for stochastic variability. Each run produces a final candidate expression together with its associated squared error. This procedure yields a set of candidate expressions, from which dominant analytic forms can emerge. The number of repetitions is chosen to balance capturing sufficient diversity while keeping the computational cost modest.

\vspace{.1cm}
\subsubsection{Selecting the best candidate}
To filter candidate expressions, each is reduced to a symbolic template by removing numerical coefficients. The most frequent template is selected as the functional form, with ties decided in favour of simpler expressions. Among candidates matching this template, the coefficients associated with the lowest squared error are retained. The procedure is applied separately to the diffusion and growth expressions, yielding the final analytic forms $D_{\mathrm{SR}}$ and $G_{\mathrm{SR}}$.

%% file: Methods/Tables/SR.tex
\begin{table}[b]
\caption{Nondefault \texttt{PySR} parameters used for symbolic regression}
\label{tab:Methods:secSR:PySR_params}
\centering
\scriptsize
\setlength{\tabcolsep}{2.8pt}
\renewcommand{\arraystretch}{1.2}
\begin{tabular}{ll>{\footnotesize}p{4cm}}
\hline
\textbf{Parameter} & \textbf{Value} & \textbf{\scriptsize Description} \\
\hline

\texttt{unary\_operators} & $\{\mathrm{sqrt}, \exp, \log\}$ &
Permitted variable transformations.
\\

\texttt{binary\_operators} & $\{+, -, \times, /\}$ &
Permitted operators for combining variables and  sub-expressions.
\\

\texttt{population\_size} & 50 &
Number of expressions retained and evolved across iterations.
\\

\texttt{niterations} & 25 &
Number of evolutionary iterations.
\\

\texttt{maxsize} & 10 &
Maximum number of operators, variables and sub-expressions.\\

\texttt{loss\_function} & \texttt{$(x - y)^2$} &
Squared error; 
output $x$, target $y$.

\\
\hline
\end{tabular}
\end{table}

%% file: Methods/figs/fig_results_schematic.tex
\begin{figure*}[t]
\centering
\begin{minipage}[t]{1\textwidth}
    \centering
    \includegraphics[width=\linewidth]{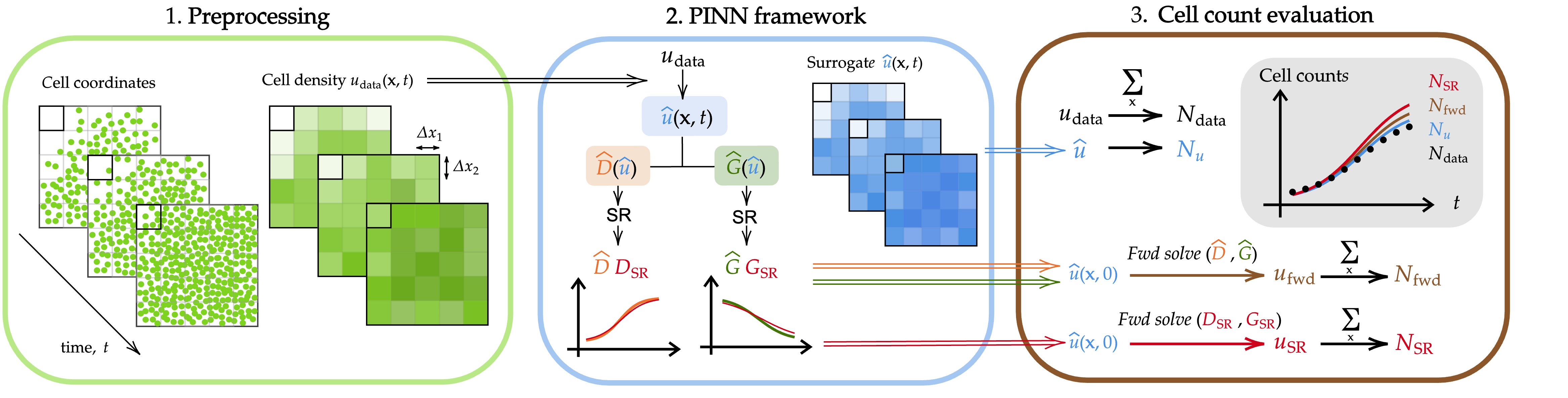}
\end{minipage}
\caption{Pipeline for applying the PINN framework on experimental microscopy data.}
\label{fig:experi_schematic}
\end{figure*}

%% file: Methods/experimental_data_pipeline.tex
\subsection{Experimental pipeline}
To apply the PINN framework to experimental data, we embed it within a three-stage pipeline: (I) preprocessing, where discrete cell coordinates \((x_1, x_2, t)\) are used to estimate the cell density field; (II) implementation of the PINN framework, 
where diffusion and growth dynamics are learned, and SR extracts interpretable functional forms; and (III) cell count evaluation, 
where total cell count dynamics resulting from step (II) are compared to the biological data. 
This workflow is shown in Fig.~\ref{fig:experi_schematic}.

\vspace{0.1cm}
\subsubsection{Preprocessing} 
We apply the pipeline to three replicate time-lapse microscopy datasets of PC9 lung cancer cells imaged every 4 hours for up to 2.5 days. 
Raw cell coordinates are converted into spatial density fields $u_{\text{data}}(\mathbf{x},t)$ via binning, 
where bin sizes are chosen to balance two considerations: ({$a$}) the spatial scales over which cells sense their local environment and regulate movement and proliferation, and ({$b$}) the need for sufficiently many cells per bin so that the resulting field behaves as a density rather than being dominated by microscopic stochasticity.
After tuning for these considerations, we select bin-sizes $\Delta x_1, \Delta x_2 \approx 0.1$~mm.  
Preprocessing yields three datasets corresponding to replicates 1–3: 
$u_{\text{data},1} \in \mathbb{R}^{15 \times 11 \times 9}$,
$u_{\text{data},2} \in \mathbb{R}^{15 \times 11 \times 12}$,
and
$u_{\text{data},3}  \in \mathbb{R}^{15 \times 11 \times 16}$, respectively.

\vspace{0.1cm}
\subsubsection{The PINN framework}
The BINN architecture is trained on each dataset independently, with all hyperparameters fixed except for the ES patience, which is varied over $\mathit{ES}\in\{500,1000,2000\}$. For each ES patience, training is performed across five TV splits. A preferred ES is then selected, defining the corresponding diffusion and growth ensembles. SR is subsequently applied ten times to these ensemble predictions, with one final functional form for $D_{\mathrm{SR}}$ and $G_{\mathrm{SR}}$ selected per replicate.

\vspace{0.1cm}
\subsubsection{Cell count evaluation}
To assess model validity, we compute total cell-count curves by summing predicted densities over space, and comparing them to the observed data count curve \( N_{\text{data}}(t) \). Predicted densities are obtained in three ways: ($i$) directly from the trained density MLP; ($ii$) by forward solving \eqref{eq:reaction_diff_eq_2d} with no flux boundary conditions using the learned diffusion and growth MLPs; and ($iii$) by forward solving \eqref{eq:reaction_diff_eq_2d} with the SR pair \( (D_{\mathrm{SR}}, G_{\mathrm{SR}}) \), with the initial condition given by the density MLP evaluated at $t=0$. The total cell counts obtained from the densities in ($i$), ($ii$), and ($iii$) are denoted by \(N_u(t)\), \(N_{\mathrm{fwd}}(t)\), and \(N_{\mathrm{SR}}(t)\), respectively.

%% file: Results/results.tex
\subsection{Learning the density} 
We first examine the agreement between the density learned by the PINN framework and the observed data. 
Despite the stochasticity introduced by binning under low cell coverage, the density data retain population-scale spatial structure that can be exploited to infer the underlying dynamics. In particular, the initial conditions exhibit heterogeneous patterns, with clustered high-density regions alongside areas nearly devoid of cells (as seen in Fig.~\ref{fig:density_preds}a). Over time, these gaps are progressively filled through the combined effects of diffusion and growth.

The learned densities across time $\hat{u}(\mathbf{x}, t)$ smooth the empirical observations while preserving the dominant spatial structure. This effect is illustrated in Fig.~\ref{fig:density_preds}b for the first replicate, shown for a fixed $x_2$ cross-section at the initial and final time points.  
Variability in the density predictions is observed across TV splits reflecting sensitivity to the specific data partitioning. This variability is expected given the high level of stochasticity in the data.

\input{Results/figs/fig_results_u}

\subsection{Learning the PDE terms}

We next examine the learned diffusion and growth functions for the three experimental replicates. As the density distributions across replicates have comparable support (Fig.~\ref{fig:main_results}a), the outputs of the independently trained diffusion and growth predictions can be meaningfully compared.

\subsubsection{Diffusion}  
All learned diffusion functions increase monotonically with increasing cell density (Fig.~\ref{fig:main_results}b). This occurs despite no monotonicity constraint being imposed during training, suggesting 
shared underlying biological behaviour. The function magnitudes are 
on the order of $10^{-2}$ mm$^2$/day 
at low densities, increasing approximately twofold (replicate 2) or tenfold (replicates 1 and 3) at high densities. This discrepancy likely reflects differences in initial conditions: replicate 2 reaches high plate coverage earlier than the other 
replicates, thereby limiting high diffusion rates (Appendix~\ref{appendix2}). 

Inspecting the SR results reveals a shared exponential structure across all three replicates. 
Specifically, replicates 1 and 2 share the template $C_0{+}C_1 e^{C_2 U}$, where $U = u/u_{\max}$ is the per-replicate normalised density, with $u_{\max}$ denoting the maximum density observed for that replicate, and $C_0, C_1, C_2{>}0$ (Table~\ref{tab:SR_main}). 
Replicate 3 exhibits a slight modification, $C_0{+}C_1 U^2 e^{C_2 U}$, introducing a multiplicative factor $U^2$ while retaining the same exponential form.

\input{Results/figs/fig_results_DG}

\subsubsection{Growth}  
Across all three replicates, the learned growth functions display a clear monotonic decrease over the central 90\% of their respective density supports (Fig.~\ref{fig:main_results}c), again without any imposed monotonicity or bounding constraints. Even more so than diffusion, the learned growth functions exhibit strong agreement both across replicates and within replicate TV splits (Fig.~\ref{fig:main_results}c). 
This suggests that, in this EQL setting, growth is more readily identifiable from data than diffusion and correspondingly plays a more dominant role in the learned dynamics.

Across the examined density range,  replicates 1 and 2 exhibit similar magnitudes, while replicate 3 shows slightly lower values.
Inspecting the SR results exposes a predominantly linear functional structure across all three replicates. 
Replicates 2 and 3 follow an identical linear form $C_0{-}C_1 U$, with $C_0,C_1{>}0$, while replicate 1 introduces a mild nonlinear correction $C_0{-}C_1 U{-}C_2 U^{3/2}$, with $C_0, C_1, C_2{>}0$.

\input{Results/tables/SR_counts_score}
\input{Results/fwd} \label{sec:fwd}
\input{Results/learning_process}

%% file: Results/figs/fig_results_u.tex
\begin{figure}[b]
    \centering
    \begin{minipage}[t]{0.24\textwidth}
        \centering
        \cornerlabelimgshift[width=\linewidth]{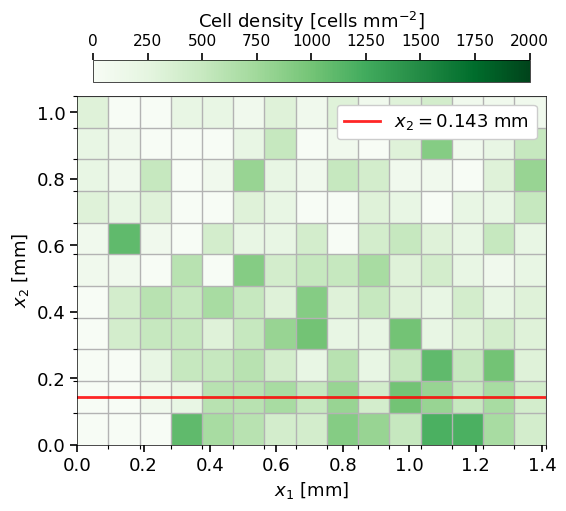}{(a)}{north}{xshift=-37pt,yshift=-23pt}
    \end{minipage}
    \hfill
    \begin{minipage}[t]{0.24\textwidth}
        \centering
        \cornerlabelimgshift[width=\linewidth]{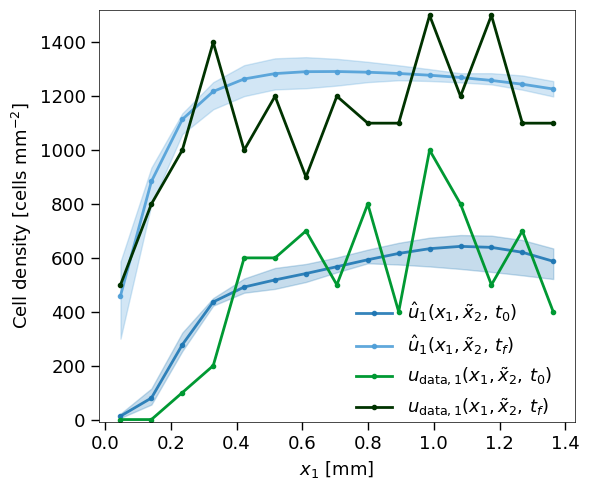}{(b)}{north}{xshift=-33pt,yshift=-4pt}
    \end{minipage}
\caption{Binned density data $u_{\text{data},1}$ and density predictions $\hat{u}_1$ for replicate 1. (a) Density data across both spatial dimensions at $t_0=0$. (b) Comparison of data and predictions at fixed $x_2 = \tilde{x}_2 = 0.143~\text{mm}$ for $t_0$ and $t_f= 52$~hours. The solid line indicates the mean across five training--validation splits, with the shaded region showing the min-max range.
}
    \label{fig:density_preds}
\end{figure}

%% file: Results/figs/fig_results_DG.tex
\begin{figure*}[t!]
    \centering

    \begin{minipage}[t]{0.325\textwidth}
        \centering
        \cornerlabelimgshift[width=\linewidth]{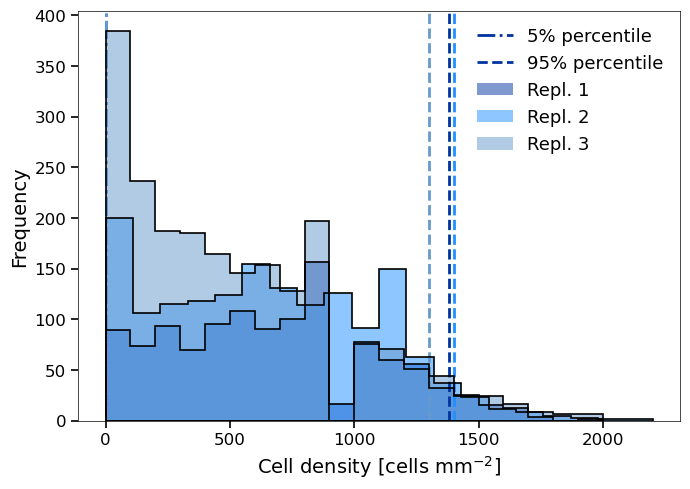}{(a)}{north}{xshift=7pt,yshift=-5pt}
    \end{minipage}
    \hfill
    \begin{minipage}[t]{0.325\textwidth}
        \centering
        \cornerlabelimgshift[width=\linewidth]{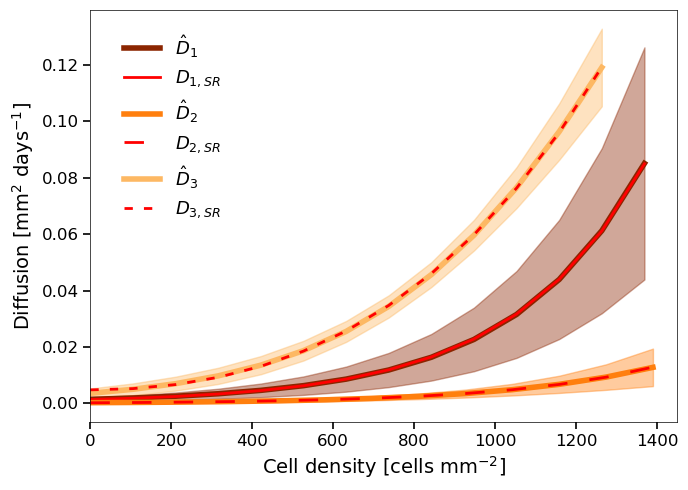}{(b)}{north}{xshift=7pt,yshift=-5pt}
    \end{minipage}
    \hfill
    \begin{minipage}[t]{0.325\textwidth}
        \centering
        \cornerlabelimgshift[width=\linewidth]{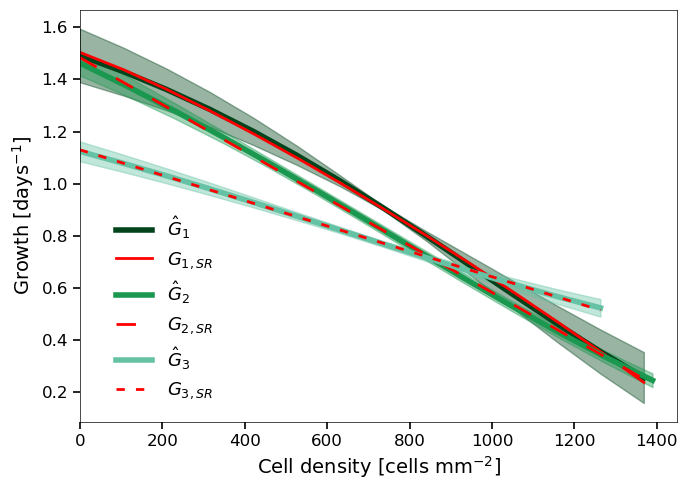}{(c)}{north}{xshift=5pt,yshift=-5pt}
    \end{minipage}

    \caption{
    Growth and diffusion predictions across five training-validation (TV) splits for each biological replicate. 
    (a) Histograms of densities used to define the central 90\% range per replicate. 
    (b)
    Diffusion predictions over the central 90\% density range. Solid lines and shaded regions denote the mean and one standard deviation across TV splits, respectively. 
    (c)
    Growth predictions, formatted as in (b).
    } 
    \label{fig:main_results}
\end{figure*}

%% file: Results/tables/SR_counts_score.tex
\begin{table}[b!]
\centering
\scriptsize
\caption{Final symbolic expressions for diffusion and growth functions}

\renewcommand{\arraystretch}{1.15}
\setlength{\tabcolsep}{3pt}

\newcommand{\tsup}[1]{\raisebox{0.6ex}{\mbox{\tiny #1}}}
\newcommand{\tsub}[1]{\raisebox{-0.6ex}{\mbox{\tiny #1}}}

\newcolumntype{Y}{>{\raggedright\arraybackslash}X}

\begin{tabularx}{\columnwidth}{@{}
    >{\centering\arraybackslash}p{0.07\columnwidth}
    >{\centering\arraybackslash}p{0.15\columnwidth}
    >{\centering\arraybackslash}p{0.15 \columnwidth}
    Y @{}}
\toprule
\textbf{Repl.} &
\shortstack{$\boldsymbol{u}$\tsub{max} \\ (cells mm\tsup{$-2$})} &
\textbf{Units} &
\multicolumn{1}{c}{\shortstack{\textbf{Symbolic expressions} \\ ($U = u / u$\tsub{max})}} \\
\midrule

1 & 2000 &
mm\tsup{2}/day &
$D$\tsub{SR,1}$=1.110\times10^{-3}e^{6.339U}+3.384\times10^{-4}$ \\
  &       &
day\tsup{$-1$} &
$G$\tsub{SR,1}$=1.503-U-1.029U^{3/2}$ \\

\midrule

2 & 2200 &
mm\tsup{2}/day &
$D$\tsub{SR,2}$=2.699\times10^{-4}e^{6.126U}-1.016\times10^{-4}$ \\
  &       &
day\tsup{$-1$} &
$G$\tsub{SR,2}$=1.484-1.983U$ \\

\midrule

3 & 2000 &
mm\tsup{2}/day &
$D$\tsub{SR,3}$=0.1527\,U^2e^U+0.004755$ \\
  &       &
day\tsup{$-1$} &
$G$\tsub{SR,3}$=1.130-0.9744U$ \\

\bottomrule
\label{tab:SR_main}
\end{tabularx}
\end{table}

%% file: Results/fwd.tex
\subsection{Evaluating the learned equation dynamics}
To assess predictive ability of the PINN framework, the learned MLP components 
and their SR counterparts are forward solved in the governing reaction--diffusion equation and compared to the observed biological data (Fig.~\ref{fig:fwd}).
The learned density MLP, trained on the spatio-temporal density data, accurately recovers the total cell counts over time across all three replicates, despite low cell numbers and substantial experimental noise.
Forward simulations using both the MLP surrogate functions and their symbolic representations closely reproduce the observed total cell-count trajectories for replicates 1 and 2, but are less accurate for replicate 3 at later time points. 
The strong agreement between the MLP-based and symbolic forward simulations is expected, given the close fit of the analytic expressions to the diffusion and growth components learned by the MLPs (Fig.~\ref{fig:main_results}b--c).

Replicate 3 displays a total cell-count trajectory with
greater curvature than replicates 1 and 2, indicating a more challenging equation discovery scenario. 
In particular, the sharp increase in cell count between the third and penultimate time points is not 
captured by the forward simulations, leading to 
deviations from the observed data at later time points. 
This mismatch suggests 
additional dynamics not fully captured by diffusion and growth alone in \eqref{eq:reaction_diff_eq_2d}. 

\input{Results/figs/fig_results_fwd}

%% file: Results/figs/fig_results_fwd.tex
\begin{figure}[t]
    \centering
    \begin{minipage}{0.4\textwidth}
        \centering
        \cornerlabelimgshift[width=\linewidth]{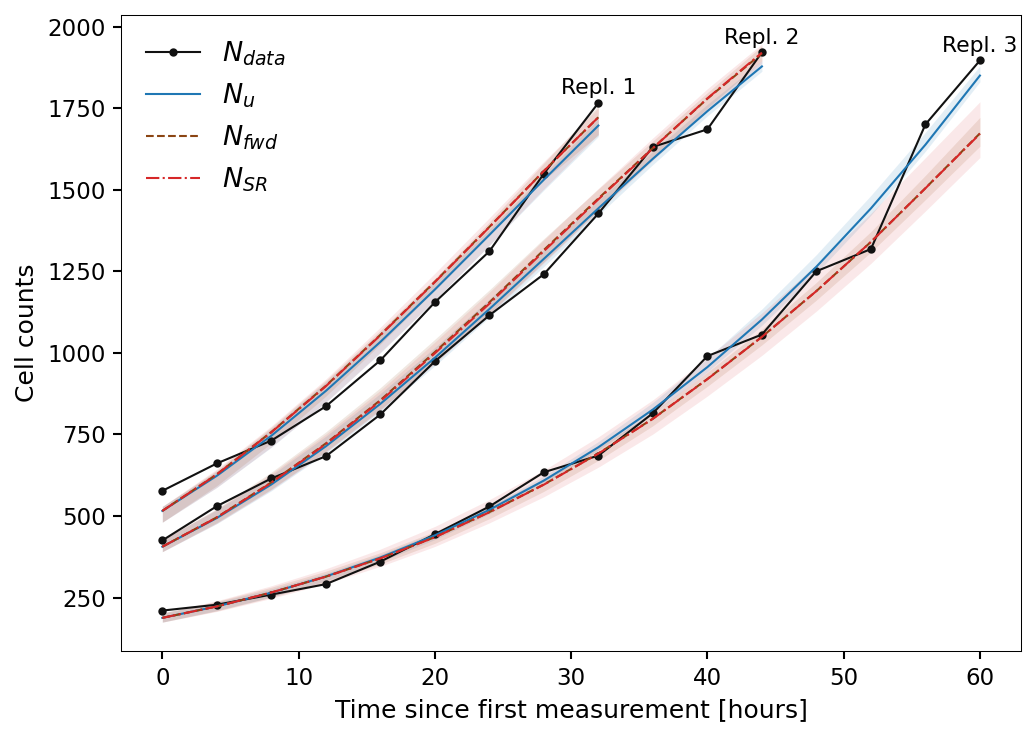}{}{north}{xshift=5pt,yshift=-5pt}
    \end{minipage}

\caption{
Total cell count trajectories. 
Observed counts $N_{\text{data}}$ are compared with total counts from the density surrogate ($N_u$), forward solutions using $\hat{D}$ and $\hat{G}$ ($N_{\text{fwd}}$), and forward solutions using $D_{\text{SR}}$ and $G_{\text{SR}}$ ($N_{\text{SR}}$). 
Solid lines show the mean, and shaded regions indicate $\pm 1$ standard deviation across training-validation splits. 
}
\label{fig:fwd}

\end{figure}

%% file: Results/learning_process.tex
\subsection{Training performance and convergence} 
To assess how the learned diffusion and growth functions emerge during training, validation loss curves and function predictions are examined across epochs (Fig. \ref{fig:experi_loss}). 
Moreover, in Table \ref{tab:final_losses_runtime}, reaction--diffusion equations obtained with ES patiences $\mathit{ES} \in \{500, 1000, 2000\}$ are compared to evaluate the trade-off between computational cost and predictive performance across all five TV splits. 
Increasing ES patience leads to substantially longer training times, with median replicate runtimes increasing from approximately $141\,\mathrm{s}$ at $\mathit{ES}=500$, to $292\,\mathrm{s}$ and $578\,\mathrm{s}$ at $\mathit{ES}=1000$ and $2000$, respectively. 
In contrast, doubling the ES patience yields only marginal improvements in validation loss, decreasing by approximately $3\%$ for both $\mathit{ES}=500\rightarrow1000$ and $1000\rightarrow2000$.
Taken together, the minimal reduction in validation loss relative to the substantial increase in runtime suggests that $\mathit{ES}=500$ is preferable. This conclusion is further supported by closer inspection of the loss components and learned functions.  

For the cell microscopy data, characterised by individual-level stochasticity,  
the validation loss decreases rapidly during early epochs as the network averages stochastic fluctuations, followed by a slower decline as larger-scale population-level structure is learned. 
This is 
observed in Fig.~\ref{fig:experi_loss}a, where the data loss 
falls quickly to approximately $2 \times 10^{-2}$ within the first few epochs, with only marginal improvements beyond $\mathit{ES}=500$. Critically, while the total validation loss continues to decrease with increasing ES patience, the PDE loss component reaches a minimum and then increases (Fig.~\ref{fig:experi_loss}b), signalling a transition from physically meaningful solutions to overfitting. 
This behaviour is reflected more strongly in the learned diffusion function, which is progressively driven toward unrealistically low values at larger epochs (Fig.~\ref{fig:experi_loss}c), whereas the learned growth function $G$ remains comparatively stable (Fig.~\ref{fig:experi_loss}d).

\input{Results/figs/fig_results_performance}

%% file: Results/figs/fig_results_performance.tex
\begin{figure}[t]
    \centering

    \begin{minipage}[t]{0.24\textwidth}
        \centering
        \cornerlabelimgshift[width=\linewidth]{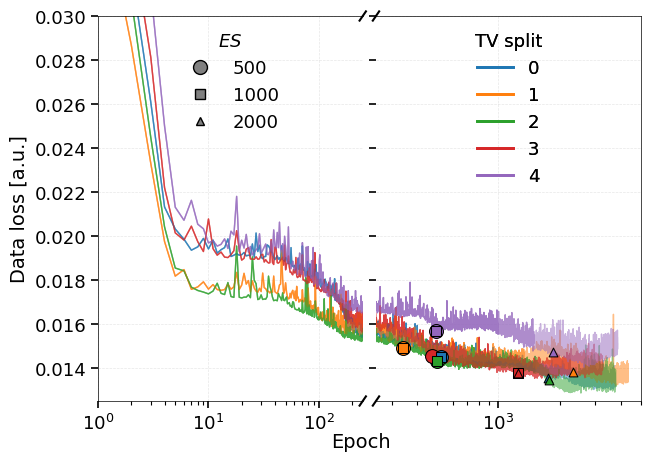}{(a)}{north}{xshift=5pt,yshift=-5pt}
    \end{minipage}
    \hfill
    \begin{minipage}[b]{0.24\textwidth}
        \centering
        \cornerlabelimgshift[width=\linewidth]{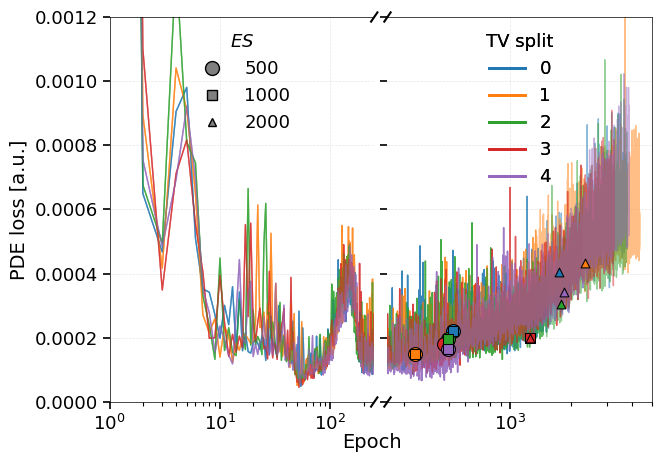}{(b)}{north}{xshift=5pt,yshift=-5pt}
    \end{minipage}

    \begin{minipage}[t]{0.24\textwidth}
        \centering
        \cornerlabelimgshift[width=\linewidth]{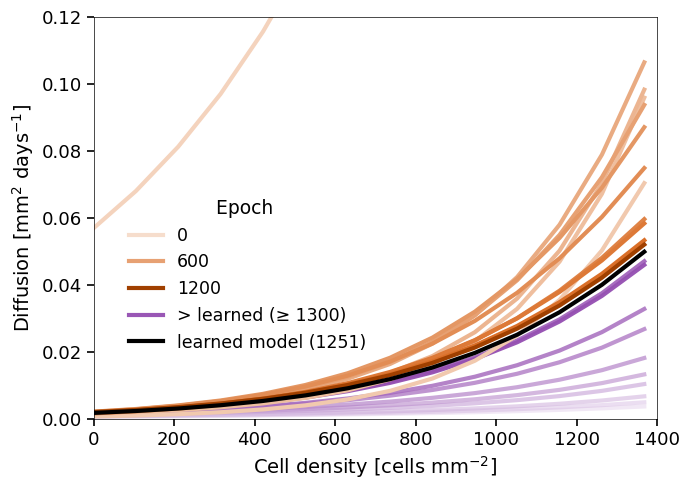}{(c)}{north}{xshift=5pt,yshift=-5pt}
    \end{minipage}
    \hfill
    \begin{minipage}[t]{0.24\textwidth}
        \centering
        \cornerlabelimgshift[width=\linewidth]{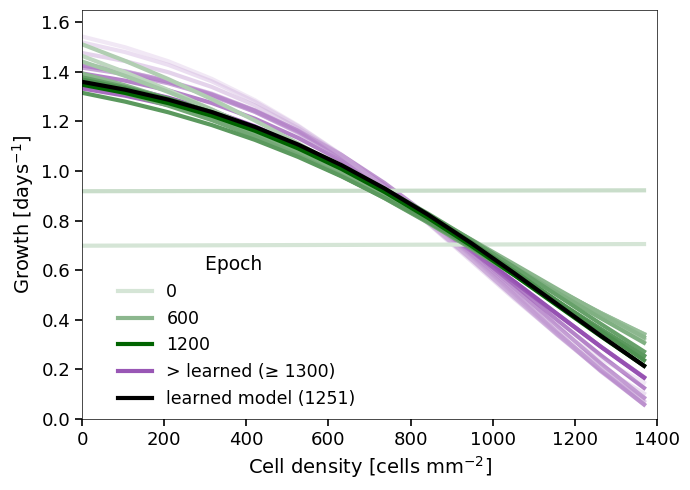}{(d)}{north}{xshift=5pt,yshift=-5pt}
    \end{minipage}

    \caption{
BINN training curves for replicate 1.
(a) Data loss history. Trained architectures for each considered early stopping (ES) patience is indicated by markers.
(b) As in (a) for PDE loss history. 
(c) Diffusion-function predictions across training epochs 
shown over the central 90\% of the density support. 
Although $\mathit{ES}{=}500$ performs best overall,
training-validation (TV) split 3 and $\mathit{ES}{=} 1000$ 
here illustrates the transition to non-physical diffusion and growth predictions at later epochs. 
Over epochs, orange curves darken toward the learned model (black), while purple curves become progressively lighter beyond the epoch when training is stopped.  
(d) As in (c), but for the growth function, with green curves replacing orange.
}
\label{fig:experi_loss}
\end{figure}

\begin{table}[t]
    \centering
    \caption{
    Validation loss and learning time ($\mathrm{mean}~[\textit{min},\textit{max}]$ across five training-validation splits)
    }
    \label{tab:experi_loss_stats}
    \scriptsize
    \renewcommand{\arraystretch}{0.9}
    \begin{tabular}{cccc}
    \toprule
    \textbf{Replicate} & \textbf{$\boldsymbol{\mathit{ES}}$ (epochs)} & \textbf{Validation Loss (a.u.)} & \textbf{Learning time (s)} \\
    \midrule

    \textbf{1}
        & 500  & 0.01497 \,\textit{[0.01449, 0.01585]} & 109.5 \,\textit{[101.9, 117.0]} \\
        & 1000 & 0.01482 \,\textit{[0.01399, 0.01585]} & 185.0 \,\textit{[148.1, 221.9]} \\
        & 2000 & 0.01420 \,\textit{[0.01375, 0.01506]} & 427.2 \,\textit{[388.5, 465.9]} \\
    \midrule

    \textbf{2}
        & 500  & 0.01305 \,\textit{[0.01248, 0.01337]} & 141.0 \,\textit{[118.4, 163.6]} \\
        & 1000 & 0.01267 \,\textit{[0.01122, 0.01337]} & 292.3 \,\textit{[177.7, 406.9]} \\
        & 2000 & 0.01225 \,\textit{[0.01122, 0.01266]} & 578.3 \,\textit{[517.0, 639.7]} \\
    \midrule

    \textbf{3}
        & 500  & 0.01257 \,\textit{[0.01199, 0.01341]} & 245.9 \,\textit{[165.8, 326.0]} \\
        & 1000 & 0.01257 \,\textit{[0.01199, 0.01341]} & 309.1 \,\textit{[300.1, 318.2]} \\
        & 2000 & 0.01205 \,\textit{[0.01146, 0.01274]} & 738.9 \,\textit{[531.0, 946.9]} \\

    \bottomrule
    \end{tabular}
    \label{tab:final_losses_runtime}
\end{table}

%% file: future_work.tex
This work presents a framework for learning reaction--diffusion dynamics from spatio-temporal density data using neural networks, specifically BINNs. 
It extends the BINN architecture previously developed by Lagergren et al.~\cite{Lagergren2020} to $2\text{D}{+}t$ and augments it by use of a symbolic regression post-processing module, forming a framework capable of discovering closed-form, interpretable governing equations directly. 
When applied to experimental cell microscopy data, the framework yields interpretable reaction--diffusion equations that reproduce observed cell population dynamics. 
These results demonstrate the potential of data-driven approaches for discovering equations governing population dynamics from data, even in biological settings where substantial stochasticity arises from individual behaviour, alongside experimental noise. 
Moreover, the proposed framework is computationally efficient and feasible for EQL workflows: a complete pipeline run for a single replicate aggregated over five TV splits and ten SR runs requires approximately 10 minutes on a standard laptop (Apple M3 Pro; CPU-only execution). 
A limitation of the present framework is that diffusion dynamics were less robustly identified than growth, as diffusion is difficult to infer when growth dominates and high cell coverage on the experimental plates limits spatial redistribution. 
Future work will investigate experimental designs that enhance diffusion identifiability, provide more principled uncertainty quantification, and extend the framework to more complex biological dynamics. 
These directions will further broaden the applicability of data-driven equation discovery, extending the framework presented here.

%% file: Appendices/A1.tex
All candidate expressions obtained from the ten \texttt{PySR} runs across all replicates are provided in Table \ref{tab:sr_combined_ieee}. The counts in each replicate column sum to 10. The expression with the highest count (out of ten) defines the dominant symbolic template for each replicate. Coefficients $C_{(\cdot)} \in \mathbb{R}$ are non-zero constants.
\input{Results/tables/SR_counts_score_template} 

%% file: Results/tables/SR_counts_score_template.tex
\begin{table}[t]
\centering
\scriptsize
\caption{All diffusion and growth expressions with bolded entries indicating the dominant template per replicate}
\label{tab:sr_combined_ieee}

\renewcommand{\arraystretch}{1.15}
\setlength{\tabcolsep}{4pt}

\begin{tabular}{>{\centering\arraybackslash} p{1.2cm}  p{3.4cm} >{\centering\arraybackslash}p{0.8cm} >{\centering\arraybackslash}p{0.8cm} >{\centering\arraybackslash}p{0.8cm} >{\centering\arraybackslash}p{0.5cm}}
\toprule
& & \multicolumn{4}{c}{\textbf{Count}} \\
\cmidrule(lr){3-6}
\raisebox{0.3ex}{\textbf{Target}} & \raisebox{0.3ex}{\textbf{Symbolic Template}} & \textbf{Repl. 1} & \textbf{Repl. 2} & \textbf{Repl. 3} & \textbf{Total} \\
\midrule
& (A) $C_0 + C_1 e^{C_2 U}$ & \textbf{5} & \textbf{6} & 0 & \textbf{11} \\
& (B) $C_0 + C_1 U^2 e^{U}$ & 0 & 0 & \textbf{5} & \textbf{5} \\
& (C) $C_0 U e^{C_1 U}$ & 0 & 2 & 1 & 3 \\
& (D) $C_0 e^{C_1 U}$ & 3 & 0 & 0 & 3 \\
\smash{\parbox[c][0pt][c]{1.2cm}{\centering \textbf{Diffusion}\\ $D_{\text{SR}}(U)$}}
& (E) $(C_0 + C_1 U)e^{C_2 U}$ & 0 & 2 & 1 & 3 \\
& (F) $C_0 e^{C_1 U + C_2 U^{1/2}}$ & 1 & 0 & 0 & 1 \\
& (G) $(C_0 U^2)/(C_1 + C_2 U)$ & 1 & 0 & 0 & 1 \\
& (H) $C_0 U(C_1 + e^{C_2 U})$ & 0 & 0 & 1 & 1 \\
& (I) $C_0 e^{C_1 U^2} + C_2$ & 0 & 0 & 1 & 1 \\
& (J) $C_0 U^{5/2} + C_1$ & 0 & 0 & 1 & 1 \\
\midrule

& (A) $C_0 + C_1 U$ & 0 & \textbf{7} & \textbf{5} & \textbf{12} \\
& (B) $C_0 + C_1 U + C_2 U^{3/2}$ & \textbf{7} & 0 & 0 & \textbf{7} \\
& (C) $C_0 e^{C_1 U e^{e^U}}$ & 0 & 2 & 0 & 2 \\
& (D) $C_0 (C_1 + C_2 U + C_3 U^3)^{1/2}$ & 0 & 0 & 2 & 2 \\
& (E) $C_0 + (C_1 + C_2 U)e^{C_3 U}$ & 0 & 1 & 0 & 1 \\
\smash{\parbox[c][0pt][c]{1.2cm}{\centering \textbf{Growth}\\ $G_{\text{SR}}(U)$}}
& (F) $C_0 + C_1 e^{C_2 U^{3/2}}$ & 1 & 0 & 0 & 1 \\
& (G) $C_0 + C_1 e^{C_2 U^2} + C_3 U$ & 1 & 0 & 0 & 1 \\
& (H) $C_0 + C_1 U^{5/4}$ & 1 & 0 & 0 & 1 \\
& (I) $C_0 + C_1 e^{C_2 U^2 + C_3 U}$ & 0 & 0 & 1 & 1 \\
& (J) $C_0 e^{C_1 + C_2 U + C_3 U^2}$ & 0 & 0 & 1 & 1 \\
& (K) $C_0 e^{C_1 U^{3/2} + C_2 U}$ & 0 & 0 & 1 & 1 \\

\bottomrule
\end{tabular}
\end{table}

%% file: Appendices/A2.tex
Fig.~\ref{fig:3x2_grid} displays spatio-temporal cell density data from the microscopy images. 
For each replicate, $\tau$ is defined as the first time point at which at least one bin exceeds a cell density of $1000$ mm$^{2}$/day. Here, $\tau = 24$ h for all three replicates.
\input{Appendices/fig/fig}

%% file: Appendices/fig/fig.tex
\begin{figure}[h]
    \centering

    \noindent
    \begin{minipage}{0.05\linewidth}
    \end{minipage}\hfill
    \begin{minipage}{0.88\linewidth}
        \centering
        \begin{minipage}{0.49\linewidth}
            \centering
             {\footnotesize \textbf{t = $\boldsymbol{\tau}$}}
        \end{minipage}\hfill
        \begin{minipage}{0.49\linewidth}
            \centering
             {\footnotesize \textbf{t = $\boldsymbol{\tau} \mathbf{+ 24}$ h}}
        \end{minipage}
    \end{minipage}

    \vspace{0.2em}

    \noindent
    \begin{minipage}{0.05\linewidth}
        \centering
        \rotatebox{90}{\footnotesize \textbf{Repl. 1}}
    \end{minipage}
    \begin{minipage}{0.88\linewidth}
        \centering
        \begin{minipage}{0.49\linewidth}
            \centering
            \cornerlabelimgshift[width=\linewidth]
            {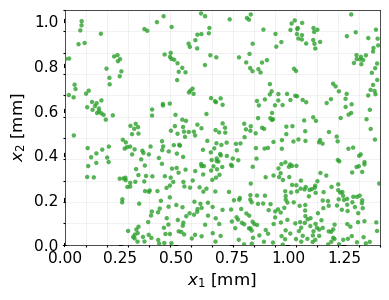}
            {}{north}{xshift=3pt,yshift=-3pt}
        \end{minipage}\hfill
        \begin{minipage}{0.49\linewidth}
            \centering
            \cornerlabelimgshift[width=\linewidth]
            {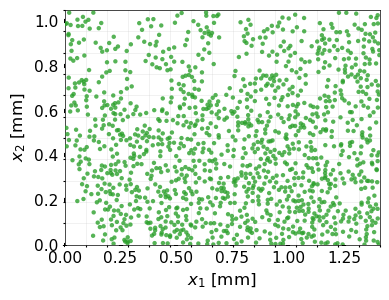}
            {}{north}{xshift=3pt,yshift=-3pt}
        \end{minipage}
    \end{minipage}

    \vspace{0.2em}

    \noindent
    \begin{minipage}{0.05\linewidth}
        \centering
        \rotatebox{90}{\footnotesize \textbf{Repl. 2}}
    \end{minipage}
    \begin{minipage}{0.88\linewidth}
        \centering
        \begin{minipage}{0.49\linewidth}
            \centering
            \cornerlabelimgshift[width=\linewidth]
            {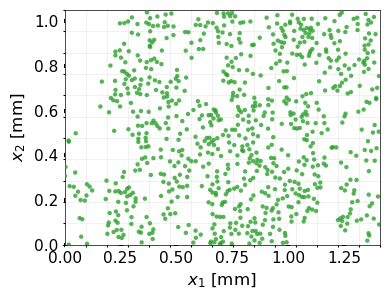}
            {}{north}{xshift=3pt,yshift=-3pt}
        \end{minipage}\hfill
        \begin{minipage}{0.49\linewidth}
            \centering
            \cornerlabelimgshift[width=\linewidth]
            {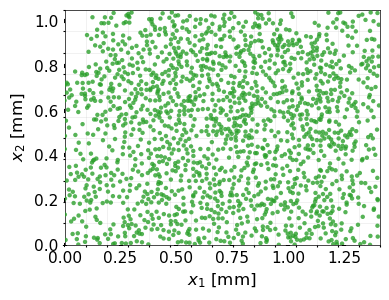}
            {}{north}{xshift=3pt,yshift=-3pt}
        \end{minipage}
    \end{minipage}

    \vspace{0.2em}

    \noindent
    \begin{minipage}{0.05\linewidth}
        \centering
        \rotatebox{90}{\footnotesize \textbf{Repl. 3}}
    \end{minipage}
    \begin{minipage}{0.88\linewidth}
        \centering
        \begin{minipage}{0.49\linewidth}
            \centering
            \cornerlabelimgshift[width=\linewidth]
            {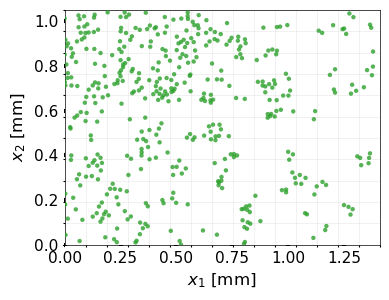}
            {}{north}{xshift=3pt,yshift=-3pt}
        \end{minipage}\hfill
        \begin{minipage}{0.49\linewidth}
            \centering
            \cornerlabelimgshift[width=\linewidth]
            {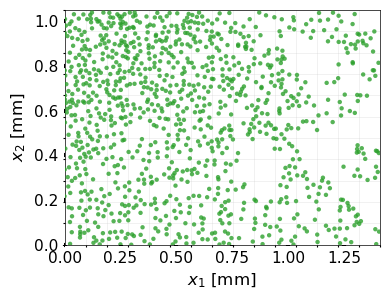}
            {}{north}{xshift=3pt,yshift=-3pt}
        \end{minipage}
    \end{minipage}

    \caption{\textcolor{black}{Microscopy images at times $\tau$ and $\tau +  24\,\mathrm{h}$ for all replicates.
    }}
    \label{fig:3x2_grid}
\end{figure}

%% file: references.bib
@article{Lagergren2020,
	title = {Biologically-informed neural networks guide mechanistic modeling from sparse experimental data},
	volume = {16},
	issn = {1553-7358},
	doi = {10.1371/journal.pcbi.1008462},
	language = {en},
	number = {12},
	urldate = {2026-04-15},
	journal = {PLoS Comput. Biol.},
	author = {Lagergren, John H. and Nardini, John T. and Baker, Ruth E. and Simpson, Matthew J. and Flores, Kevin B.},
	editor = {Lavrik, Inna},
	month = dec,
	year = {2020},
	pages = {e1008462},
}

@misc{Cranmer2023,
	title = {Interpretable {Machine} {Learning} for {Science} with {PySR} and {SymbolicRegression}.jl},
	copyright = {arXiv.org perpetual, non-exclusive license},
	url = {https://arxiv.org/abs/2305.01582},
	doi = {10.48550/ARXIV.2305.01582},
	urldate = {2026-04-15},
	publisher = {arXiv},
	author = {Cranmer, Miles},
	year = {2023},
  note   = {arXiv:2305.01582},
}

@article{Lagergren2020_b,
	title = {Learning partial differential equations for biological transport models from noisy spatio-temporal data},
	volume = {476},
	issn = {1364-5021, 1471-2946},
	doi = {10.1098/rspa.2019.0800},
	language = {en},
	number = {2234},
	urldate = {2026-04-15},
	journal = {Proc. R. Soc. Math. Phys. Eng. Sci.},
	author = {Lagergren, John H. and Nardini, John T. and Michael Lavigne, G. and Rutter, Erica M. and Flores, Kevin B.},
	month = feb,
	year = {2020},
	pages = {20190800},
}

@misc{Gao2025,
	title = {Mesh-free sparse identification of nonlinear dynamics},
	copyright = {arXiv.org perpetual, non-exclusive license},
    url = {https://arxiv.org/abs/2505.16058},
	doi = {10.48550/ARXIV.2505.16058},
	urldate = {2026-04-15},
	publisher = {arXiv},
	author = {Gao, Mars Liyao and Kutz, J. Nathan and Font, Bernat},
	year = {2025},
    note = {arXiv:2505.16058},
}

@article{Stephany2024,
	title = {{PDE}-{LEARN}: {Using} deep learning to discover partial differential equations from noisy, limited data},
	volume = {174},
	issn = {08936080},
	shorttitle = {{PDE}-{LEARN}},
	doi = {10.1016/j.neunet.2024.106242},
	language = {en},
	urldate = {2026-04-15},
	journal = {Neural Netw.},
	author = {Stephany, Robert and Earls, Christopher},
	month = jun,
	year = {2024},
	pages = {106242},
}

@article{Chen2021,
	title = {Physics-informed learning of governing equations from scarce data},
	volume = {12},
	issn = {2041-1723},
	doi = {10.1038/s41467-021-26434-1},
	language = {en},
	number = {1},
	urldate = {2026-04-15},
	journal = {Nat. Commun.},
	author = {Chen, Zhao and Liu, Yang and Sun, Hao},
	month = oct,
	year = {2021},
	pages = {6136},
}

@misc{rackauckas2021universaldifferentialequationsscientific,
	title = {Universal {Differential} {Equations} for {Scientific} {Machine} {Learning}},
	copyright = {arXiv.org perpetual, non-exclusive license},
	url = {https://arxiv.org/abs/2001.04385},
	doi = {10.48550/ARXIV.2001.04385},
	urldate = {2026-04-15},
	publisher = {arXiv},
	author = {Rackauckas, Christopher and Ma, Yingbo and Martensen, Julius and Warner, Collin and Zubov, Kirill and Supekar, Rohit and Skinner, Dominic and Ramadhan, Ali and Edelman, Alan},
	year = {2020},
	note = {arXiv:2001.04385},
}

@inproceedings{Podina2024UPINN,
  title = 	 {Universal Physics-Informed Neural Networks: Symbolic Differential Operator Discovery with Sparse Data},
  author =       {Podina, Lena and Eastman, Brydon and Kohandel, Mohammad},
  booktitle = 	 {Proceedings of the 40th International Conference on Machine Learning},
  pages = 	 {27948--27956},
  year = 	 {2023},
  editor = 	 {Krause, Andreas and Brunskill, Emma and Cho, Kyunghyun and Engelhardt, Barbara and Sabato, Sivan and Scarlett, Jonathan},
  volume = 	 {202},
  series = 	 {Proceedings of Machine Learning Research},
  month = 	 {23--29 Jul},
  publisher =    {PMLR},
}

@article{Li2024ops,
	title = {Physics-{Informed} {Neural} {Operator} for {Learning} {Partial} {Differential} {Equations}},
	volume = {1},
	copyright = {https://www.acm.org/publications/policies/copyright\_policy\#Background},
	issn = {2831-3194},
	doi = {10.1145/3648506},
	language = {en},
	number = {3},
	urldate = {2026-04-15},
	journal = {JDS},
	author = {Li, Zongyi and Zheng, Hongkai and Kovachki, Nikola and Jin, David and Chen, Haoxuan and Liu, Burigede and Azizzadenesheli, Kamyar and Anandkumar, Anima},
	month = sep,
	year = {2024},
	pages = {1--27},
}

@article{Liu2024,
	title = {Multi-resolution partial differential equations preserved learning framework for spatiotemporal dynamics},
	volume = {7},
	issn = {2399-3650},
	doi = {10.1038/s42005-024-01521-z},
	language = {en},
	number = {1},
	urldate = {2026-04-15},
	journal = {Commun. Phys.},
	author = {Liu, Xin-Yang and Zhu, Min and Lu, Lu and Sun, Hao and Wang, Jian-Xun},
	month = jan,
	year = {2024},
	pages = {31},
}

@article{Liu2021,
	title = {B-{PINNs}: {Bayesian} physics-informed neural networks for forward and inverse {PDE} problems with noisy data},
	volume = {425},
	issn = {00219991},
	shorttitle = {B-{PINNs}},
	doi = {10.1016/j.jcp.2020.109913},
	language = {en},
	urldate = {2026-04-15},
	journal = {J. Comput. Phys.},
	author = {Yang, Liu and Meng, Xuhui and Karniadakis, George Em},
	month = jan,
	year = {2021},
	pages = {109913},
}

@article{Raissi2019,
	title = {Physics-informed neural networks: {A} deep learning framework for solving forward and inverse problems involving nonlinear partial differential equations},
	volume = {378},
	issn = {00219991},
	shorttitle = {Physics-informed neural networks},
	doi = {10.1016/j.jcp.2018.10.045},
	language = {en},
	urldate = {2026-04-15},
	journal = {J. Comput. Phys.},
	author = {Raissi, M. and Perdikaris, P. and Karniadakis, G. E.},
	month = feb,
	year = {2019},
	pages = {686--707},
}

@article{Becsei2024,
	title = {Time-series sewage metagenomics distinguishes seasonal, human-derived and environmental microbial communities potentially allowing source-attributed surveillance},
	volume = {15},
	issn = {2041-1723},
	doi = {10.1038/s41467-024-51957-8},
	language = {en},
	number = {1},
	urldate = {2026-04-15},
	journal = {Nat. Commun.},
	author = {Becsei, {\'A}gnes and Fuschi, Alessandro and Otani, Saria and Kant, Ravi and Weinstein, Ilja and Alba, Patricia and Stéger, József and Visontai, Dávid and Brinch, Christian and De Graaf, Miranda and Schapendonk, Claudia M. E. and Battisti, Antonio and De Cesare, Alessandra and Oliveri, Chiara and Troja, Fulvia and Sironen, Tarja and Vapalahti, Olli and Pasquali, Frédérique and Bányai, Krisztián and Makó, Magdolna and Pollner, Péter and Merlotti, Alessandra and Koopmans, Marion and Csabai, Istvan and Remondini, Daniel and Aarestrup, Frank M. and Munk, Patrick},
	month = aug,
	year = {2024},
	pages = {7551},
}

@article{Spinelli2025,
	title = {Wearable microfluidic biosensors with haptic feedback for continuous monitoring of hydration biomarkers in workers},
	volume = {8},
	issn = {2398-6352},
	doi = {10.1038/s41746-025-01466-9},
	language = {en},
	number = {1},
	urldate = {2026-04-15},
	journal = {NPJ Digit. Med.},
	author = {Spinelli, Julia C. and Suleski, Brandon J. and Wright, Donald E. and Grow, Joseph L. and Fagans, Gabriel R. and Buckley, Maura J. and Yang, Da Som and Yang, Kaitao and Beil, Steven M. and Wallace, Jessica C. and DiZoglio, Thomas S. and Model, Jeffrey B. and Love, Shirley and Macintosh, David E. and Scarth, Alan P. and Marrapode, Matthew T. and Serviente, Corinna and Avila, Raudel and Alahmad, Barrak K. and Busa, Michael A. and Wright, John A. and Li, Weihua and Casa, Douglas J. and Rogers, John A. and Lee, Stephen P. and Ghaffari, Roozbeh and Aranyosi, Alexander J.},
	month = feb,
	year = {2025},
	pages = {76},
}

@article{Clausen2025,
	title = {Wearable continuous diffusion-based skin gas analysis},
	volume = {16},
	issn = {2041-1723},
	doi = {10.1038/s41467-025-59629-x},
	language = {en},
	number = {1},
	urldate = {2026-04-15},
	journal = {Nat. Commun.},
	author = {Clausen, David and Farley, Max and Little, Abigail and Kasper, Kevin and Moreno, Joseph and Limesand, Larissa and Gutruf, Philipp},
	month = may,
	year = {2025},
	pages = {4343},
}

@article{Wang2025,
	title = {Label-free evaluation of mouse embryo quality using time-lapse bright field and optical coherence microscopy},
	volume = {8},
	issn = {2399-3642},
	doi = {10.1038/s42003-025-08044-5},
	language = {en},
	number = {1},
	urldate = {2026-04-15},
	journal = {Commun. Biol},
	author = {Wang, Fei and Hao, Senyue and Park, Kibeom and Ahmady, Ali and Zhou, Chao},
	month = apr,
	year = {2025},
	pages = {612},
}

@article{Yu2025,
	title = {Longitudinal single-cell multiomic atlas of high-risk neuroblastoma reveals chemotherapy-induced tumor microenvironment rewiring},
	volume = {57},
	issn = {1546-1718},
	doi = {10.1038/s41588-025-02158-6},
	number = {5},
	journal = {Nat. Genet},
	author = {Yu, Wenbao and Biyik-Sit, Rumeysa and Uzun, Yasin and Chen, Chia-Hui and Thadi, Anusha and Sussman, Jonathan H. and Pang, Minxing and Wu, Chi-Yun and Grossmann, Liron D. and Gao, Peng and Wu, David W. and Yousey, Aliza and Zhang, Mei and Turn, Christina S. and Zhang, Zhan and Bandyopadhyay, Shovik and Huang, Jeffrey and Patel, Tasleema and Chen, Changya and Martinez, Daniel and Surrey, Lea F. and Hogarty, Michael D. and Bernt, Kathrin and Zhang, Nancy R. and Maris, John M. and Tan, Kai},
	month = may,
	year = {2025},
	pages = {1142--1154},
}

@inproceedings{neuralODE,
 author = {Chen, Ricky T. Q. and Rubanova, Yulia and Bettencourt, Jesse and Duvenaud, David K},
 booktitle = {Advances in Neural Information Processing Systems},
 editor = {S. Bengio and H. Wallach and H. Larochelle and K. Grauman and N. Cesa-Bianchi and R. Garnett},
 pages = {},
 publisher = {Curran Associates, Inc.},
 title = {Neural Ordinary Differential Equations},
 volume = {31},
 year = {2018}
}

@article{Champion2019,
	title = {Data-driven discovery of coordinates and governing equations},
	volume = {116},
	issn = {0027-8424, 1091-6490},
	doi = {10.1073/pnas.1906995116},
	language = {en},
	number = {45},
	urldate = {2026-04-15},
	journal = {Proc. Natl. Acad. Sci. U.S.A},
	author = {Champion, Kathleen and Lusch, Bethany and Kutz, J. Nathan and Brunton, Steven L.},
	month = nov,
	year = {2019},
	pages = {22445--22451},
}

@article{Fung2025,
	title = {Rapid {Bayesian} identification of sparse nonlinear dynamics from scarce and noisy data},
	volume = {481},
	issn = {1471-2946},
	doi = {10.1098/rspa.2024.0200},
	language = {en},
	number = {2307},
	urldate = {2026-04-15},
	journal = {Proc. R. Soc. Math. Phys. Eng. Sci.},
	author = {Fung, Lloyd and Fasel, Urban and Juniper, Matthew},
	month = feb,
	year = {2025},
	pages = {20240200},
}

@article{Rudy2017,
	title = {Data-driven discovery of partial differential equations},
	volume = {3},
	issn = {2375-2548},
	doi = {10.1126/sciadv.1602614},
	language = {en},
	number = {4},
	urldate = {2026-04-15},
	journal = {Sci. Adv.},
	author = {Rudy, Samuel H. and Brunton, Steven L. and Proctor, Joshua L. and Kutz, J. Nathan},
	month = apr,
	year = {2017},
	pages = {e1602614},
}

@article{Messenger2021WSINDyPDE,
	title = {Weak {SINDy} for partial differential equations},
	volume = {443},
	issn = {00219991},
	doi = {10.1016/j.jcp.2021.110525},
	language = {en},
	urldate = {2026-04-15},
	journal = {J. Comput. Phys.},
	author = {Messenger, Daniel A. and Bortz, David M.},
	month = oct,
	year = {2021},
	pages = {110525},
}

@article{Messenger2021WSINDy,
	title = {Weak {SINDy}: {Galerkin}-{Based} {Data}-{Driven} {Model} {Selection}},
	volume = {19},
	issn = {1540-3459, 1540-3467},
	shorttitle = {Weak {SINDy}},
	doi = {10.1137/20M1343166},
	language = {en},
	number = {3},
	urldate = {2026-04-15},
	journal = {Multiscale Model. Simul.},
	author = {Messenger, Daniel A. and Bortz, David M.},
	month = jan,
	year = {2021},
	pages = {1474--1497},
}

@article{Brunton2016,
	title = {Discovering governing equations from data by sparse identification of nonlinear dynamical systems},
	volume = {113},
	issn = {0027-8424, 1091-6490},
	doi = {10.1073/pnas.1517384113},
	language = {en},
	number = {15},
	urldate = {2026-04-15},
	journal = {Proc. Natl. Acad. Sci. U.S.A.},
	author = {Brunton, Steven L. and Proctor, Joshua L. and Kutz, J. Nathan},
	month = apr,
	year = {2016},
	pages = {3932--3937},
}

@article{Wei2022,
	title = {Sparse dynamical system identification with simultaneous structural parameters and initial condition estimation},
	volume = {165},
	issn = {09600779},
	doi = {10.1016/j.chaos.2022.112866},
	language = {en},
	urldate = {2026-04-15},
	journal = {Chaos Soliton Fract.},
	author = {Wei, Baolei},
	month = dec,
	year = {2022},
	pages = {112866},
}

@book{Brauer2019,
	series = {Texts in {Applied} {Mathematics}},
	title = {Mathematical {Models} in {Epidemiology}},
	isbn = {978-1-4939-9828-9},
	publisher = {Springer New York},
	author = {Brauer, F. and Castillo-Chavez, C. and Feng, Z.},
    address ={New York, NY},
	year = {2019},
}

@book{CornishBowden2013,
	title = {Fundamentals of {Enzyme} {Kinetics}},
	isbn = {978-1-4831-6119-8},
	publisher = {Butterworth-Heinemann},
	author = {Cornish-Bowden, A.},
    address ={London},
	year = {2014},
}

@book{Allen2007,
	title = {An {Introduction} to {Mathematical} {Biology}},
	isbn = {978-0-13-035216-3},
	publisher = {Pearson/Prentice Hall},
	author = {Allen, L. J. S.},
	year = {2007},
	lccn = {2006042585},
    address   = {Upper Saddle River, NJ},
}

@article{campsvalls2023discoveringcausalrelationsequations,
	title = {Discovering causal relations and equations from data},
	volume = {1044},
	issn = {03701573},
	doi = {10.1016/j.physrep.2023.10.005},
	language = {en},
	urldate = {2026-04-15},
	journal = {Phys. Rep.},
	author = {Camps-Valls, Gustau and Gerhardus, Andreas and Ninad, Urmi and Varando, Gherardo and Martius, Georg and Balaguer-Ballester, Emili and Vinuesa, Ricardo and Diaz, Emiliano and Zanna, Laure and Runge, Jakob},
	month = dec,
	year = {2023},
	pages = {1--68},
}

@article{Baker2018,
	title = {Mechanistic models versus machine learning, a fight worth fighting for the biological community?},
	volume = {14},
	issn = {1744-9561, 1744-957X},
	doi = {10.1098/rsbl.2017.0660},
	language = {en},
	number = {5},
	urldate = {2026-04-15},
	journal = {Biol. Lett.},
	author = {Baker, Ruth E. and Peña, Jose-Maria and Jayamohan, Jayaratnam and Jérusalem, Antoine},
	month = may,
	year = {2018},
	pages = {20170660},
}

@article{Jagtap2020,
	title = {Extended {Physics}-{Informed} {Neural} {Networks} ({XPINNs}): {A} {Generalized} {Space}-{Time} {Domain} {Decomposition} {Based} {Deep} {Learning} {Framework} for {Nonlinear} {Partial} {Differential} {Equations}},
	volume = {28},
	issn = {1815-2406, 1991-7120},
	shorttitle = {Extended {Physics}-{Informed} {Neural} {Networks} ({XPINNs})},
	doi = {10.4208/cicp.OA-2020-0164},
	number = {5},
	urldate = {2026-04-15},
	journal = {Commun. Comput. Phys.},
	author = {D. Jagtap, Ameya and Em Karniadakis, George},
	month = jun,
	year = {2020},
	pages = {2002--2041},
}

@misc{lavery2026binn,
  author = {William Lavery and others},
  title = {{H}yperparameter {S}election in {B}iologically-{I}nformed {N}eural {N}etworks},
  year = {2026},
  note = {Preprint, to appear}
}
